%% file: main.tex
\documentclass[letterpaper, 10 pt, conference]{ieeeconf}  %

\IEEEoverridecommandlockouts                              %

\usepackage{changes}
\usepackage{graphicx}
\usepackage{multicol}
\usepackage{booktabs}
\usepackage{multirow}
\usepackage{tabularx}
\newcolumntype{Y}{>{\centering\arraybackslash}X}
\newcolumntype{L}[1]{>{\raggedright\let\newline\\\arraybackslash\hspace{0pt}}m{#1}}
\newcolumntype{C}[1]{>{\centering\let\newline\\\arraybackslash\hspace{0pt}}m{#1}}
\newcolumntype{R}[1]{>{\raggedleft\let\newline\\\arraybackslash\hspace{0pt}}m{#1}}

\usepackage{dingbat}

\usepackage{pifont}
\newcommand{\xmark}{\ding{55}}%

\usepackage{hyperref}
\hypersetup{
    colorlinks=true,
    linkcolor=blue,
    filecolor=magenta,      
    urlcolor=blue,
    pdftitle={Overleaf Example},
    pdfpagemode=FullScreen,
    }
\usepackage{xcolor}

\usepackage{xspace}
\makeatletter
\DeclareRobustCommand\onedot{\futurelet\@let@token\@onedot}
\def\@onedot{\ifx\@let@token.\else.\null\fi\xspace}
\def\eg{\emph{e.g}\onedot} 
\def\ie{\emph{i.e}\onedot}

\makeatother

\newcommand{\nsap}[0]{{MuBlE}\xspace} %
\newcommand{\dataset}[0]{{SHOP-VRB2}\xspace} %

\title{\LARGE \bf
MuBlE: MuJoCo and Blender simulation Environment and Benchmark for Task Planning in Robot Manipulation}

\author{Michal Nazarczuk$^{1}$,  Karla Stepanova$^{2}$, Jan Kristof Behrens$^{2}$, Matej Hoffmann$^{2}$ and Krystian Mikolajczyk$^{3}$%
\thanks{$^{1}$Michal Nazarczuk is with Huawei Noah's Ark London
        {\tt\small michal.nazarczuk1@huawei.com}}%
\thanks{$^{2}$Jan Kristof Behrens, Karla Stepanova and Matej Hoffmann are with the Czech Technical University in Prague
        {\tt\small [jan.kristof.behrens, karla.stepanova]@cvut.cz}, }
\thanks{$^{3}$Krystian Mikolajczyk is with Imperial College London
        {\tt\small k.mikolajczyk@imperial.ac.uk}}%
\thanks{J.K.S., K.S., and M.H. were supported by the European Union under the project ROBOPROX (reg. no. CZ.02.01.01/00/22\_008/0004590). This work originated in the project Interactive Perception-Action-Learning for Modelling Objects (IPALM) (H2020-FET-ERA-NET Cofund-CHIST-ERA III / TAČR EPSILON, No. TH05020001)}
}

\begin{document}

\maketitle
\thispagestyle{empty}
\pagestyle{empty}

\input{Sections/abstract}
\input{Sections/introduction}

\input{Sections/related_work}

\input{Sections/environment}

\input{Sections/dataset}
\input{Sections/baseline}

\input{Sections/conclusions}

\bibliographystyle{IEEEtran}
\bibliography{references}

\end{document}

%% file: Sections/abstract.tex
\begin{abstract}

Current embodied reasoning agents struggle to plan for long-horizon tasks that require to physically interact with the world to obtain the necessary information {(\eg \emph{sort the objects from lightest to heaviest})}. The improvement of the capabilities of such an agent is highly dependent on the availability of relevant training environments. In order to facilitate the development of such systems, we introduce a novel simulation environment (built on top of $\mathtt{robosuite}$) that makes use of the MuJoCo physics engine and high-quality renderer Blender to provide realistic visual observations that are also accurate to the physical state of the scene. It is the first simulator focusing on long-horizon robot manipulation tasks preserving accurate physics modeling. \nsap{} can generate mutlimodal data for training and enable design of closed-loop methods through environment interaction on two levels: visual -- action loop, and control -- physics loop. Together with the simulator, we propose \dataset{}, a new benchmark composed of 10 classes of multi-step reasoning scenarios that require simultaneous visual and physical measurements. 

\end{abstract}

%% file: Sections/introduction.tex
\section{Introduction}
\label{sec:intro}

Many robotics systems include an agent that is required to perform a task towards a specified goal given visual observations or instructions in natural language \cite{Deitke2020RoboTHOR:Result, Ehsani2021ManipulaTHOR:Manipulation, Shridhar2020ALFREDTasks, Szot2021HabitatHabitat}. Such tasks require a simulation environment for generating data or on-line agent training, together with an evaluation process. Existing environments \cite{Kolve2017AI2-THOR:AI, Makoviychuk2021IsaacLearning} differ in terms of physics simulation engines and the quality of visual observations. The high computational load of realistic rendering forces a compromise between real-time physics calculation and the visual quality of the simulation. Realistic synthetic visual data is crucial for the development of robotics systems, but the collection of such data is often restricted by a cumbersome setup process and real-time robot operations \cite{Collins2020BenchmarkingDataset}. More powerful simulators, datasets, and benchmarks are needed for a closed-loop setup where changes to the world or physical measurements (\eg, weight, elasticity, occlusions, etc.) should be accounted for as directly affecting the next step in action planning. 

In this paper, we make the following contributions\footnote{The environment and the benchmark are released on the GitHub page \href{https://github.com/michaal94/MuBlE}{https://github.com/michaal94/MuBlE}}:

\begin{figure}[t]
    \centering
    \includegraphics[width=0.95\linewidth]{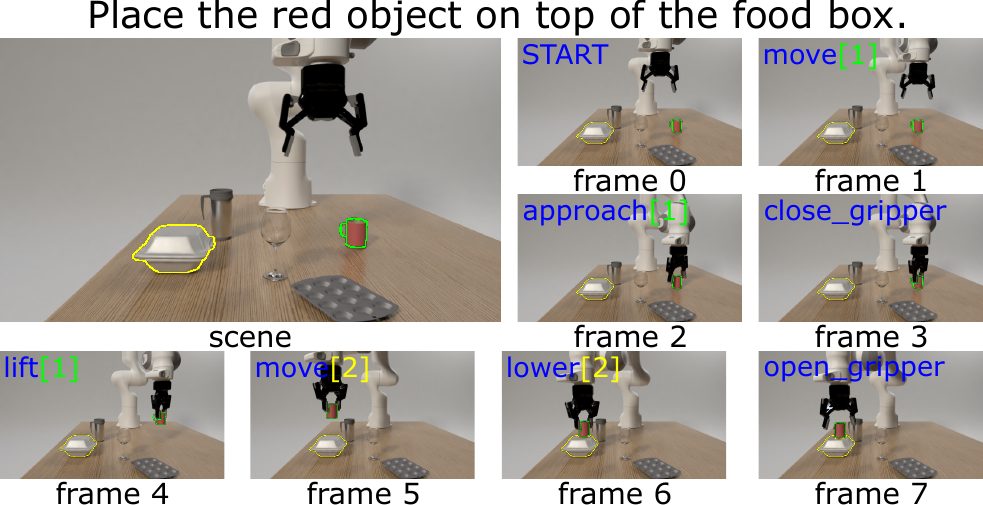}
    \caption{An example task from the proposed \dataset{} benchmark presenting capabilities of the proposed \nsap{} environment: synthetic scene and instruction generation, execution of symbolic actions for manipulation followed by physics calculation and realistic rendering. Symbolic actions with corresponding targets marked in the image.}
    \label{fig:teaser}
    \vspace{-1em}
\end{figure}
 
\begin{itemize} 
\item \nsap{}: A modular simulation environment built on $\mathtt{robosuite}$~\cite{zhu2020robosuite}, using MuJoCo~\cite{Todorov_Erez_Tassa_2012} for realistic physics and Blender~\cite{BlenderOnlineCommunity2018BlenderPackage} for high-quality rendering to enhance sim-to-real transfer. \nsap{} enables multimodal data generation, including scene and instruction synthesis, the generation of ground truth scene graphs with object attributes, task completion assessment, and a set of primitive action controllers that also support observing physical object properties (e.g., weight, stiffness). Its high-speed physics and keyframe action-render loops facilitate the integration of closed-loop reasoning methods. To our knowledge, \nsap{} is the first environment providing training data for robot manipulation task planning while ensuring both high-quality rendering and precise physics modeling.
\item \dataset{}: A benchmark for evaluating embodied, closed-loop reasoning in robot manipulation. It includes 12,000 scenes with instructions across ten types of single- and multi-step tabletop manipulation tasks, requiring reasoning over visual attributes (e.g., object properties, relations) and physical measurements (e.g., weight, stiffness). \dataset{} extends~\cite{Nazarczuk2020SHOP-VRB:Perception} with a more challenging multimodal setup, integrating vision, language, and manipulation for tasks such as Visual Question Answering (VQA)\cite{Vedantam2019ProbabilisticAnswering}, Embodied Question Answering (EQA)\cite{behrens2021embodied, Das2018EmbodiedAnswering}, and Visual-Language-Action (VLA)~\cite{kim24openvla}. It also includes 30 scenes with YCB objects~\cite{Xiang2018PoseCNN:Scenes} for sim-to-real transfer validation.
\item We demonstrate the effectiveness of our environment and benchmark by integrating them with the CLIER (Closed-Loop Interactive Embodied Reasoning)~\cite{Nazarczuk2025}, comparing simulation with real-world experiments. Our results highlight the capability of \nsap{} to bridge the gap between simulation and real-world robot manipulation.
\end{itemize}

%% file: Sections/related_work.tex
\section{Related work}
\label{sec:relwork}

\begin{table*}[]
    \centering
    \caption{\scriptsize Comparison of the \nsap{}'s capabilities with the related simulators for the robotic experiments. Our \nsap{} can be characterised by high-quality rendering with Blender, accurate physics with MuJoCo, included visible and non-visible states, various action spaces available, targeting tabletop manipulation, and being modular and easily extendable (non-kin.=non-kinematic).}
    \scriptsize
    \setlength{\tabcolsep}{1pt}
    \begin{tabular}{l c c c c c c c c c c c c}
    \toprule
         \multirow{2}[1]{*}{Simulator} & \multicolumn{3}{c}{Rendering} & \multicolumn{2}{c}{Physics} 	& \multirow{2}[1]{*}{\begin{tabular}{@{}c@{}}Motion \\ planner\end{tabular}} 	& \multirow{2}[1]{*}{\begin{tabular}{@{}c@{}}Non-kin. \\ states\end{tabular}} & \multirow{2}[1]{*}{\begin{tabular}{@{}c@{}}Non-visible \\ states\end{tabular}}	& \multirow{2}[1]{*}{Action space}	& \multirow{2}[1]{*}{Speed} & \multirow{2}[1]{*}{Scale} & \multirow{2}[1]{*}{Open-source} \\\cmidrule(lr){2-4} \cmidrule(rl){5-6}
          & library &	material & quality	& library	& supports	&  	&  & 	& 	&	&  &\\
         \hline
         AI2-THOR~\cite{Kolve2017AI2-THOR:AI}& Unity & textures & {\!++} & Unity & rigid dyn./animation& \xmark & \color{teal}{\checkmark} & \xmark & Discrete & {\!+} & room & \color{teal}{\checkmark}\\
         ManipulaTHOR~\cite{Ehsani2021ManipulaTHOR:Manipulation} & Unity &textures & {\!++} & Unity & AI2THOR+manip.& \xmark & \xmark & \xmark & Discrete & {\!+} & room & \color{teal}{\checkmark}\\
         ThreeDWorld~\cite{Gan2021ThreeDWorld:Simulation}  & Unity/V-Ray (off.)& Configurable & \color{teal}{\!+++} &Unity+FLEX &rigid/particle dyn.  & \xmark & \xmark & \xmark & Continuous & {\!++} & house & \xmark\\
        ISAACSim~\cite{Makoviychuk2021IsaacLearning} & Omniverse RTX & Configurable  &{\!++} & PhysX &rigid/artic. dyn. & \color{teal}{\checkmark} & \xmark & \xmark & (Disc.)\&Cont. & {\!++} & house  & \xmark \\
        Habitat 2.0~\cite{Szot2021HabitatHabitat} & Magnum &3D scans/PBR & {\!+} &PyBullet& rigid/artic. dyn.& \xmark & \xmark & \xmark & Continuous & {\!+++}& house & \color{teal}{\checkmark}\\
        CoppeliaSim~\cite{coppeliaSim} & OpenGL & Gouraoud shad. & {\!+} & PyBullet& rigid/artic. dyn.& \color{teal}{\checkmark} & \xmark & \xmark & (Disc.)\&Cont. & {\!++} & room & \xmark \\
        iGibson~\cite{li2022igibson} & PyRender/OpenGL & PBR shad. &{\!+} & PyBullet& rigid/artic. dyn.& \color{teal}{\checkmark} & \color{teal}{\checkmark} & \xmark & Continuous & {\!++} & house & \color{teal}{\checkmark}\\
        $\mathtt{robosuite}$~\cite{zhu2020robosuite} & MuJoCo/NVISII & Configurable &{\!++} & MuJoCo& rigid/artic. dyn.& \color{teal}{\checkmark} & \xmark & \xmark & Continuous & {\!++} & tabletop & \color{teal}{\checkmark} \\
        \hline
        {\bf MuBle (ours)} & MuJoCo/Blender & procedural &{\bf\color{teal}{\!+++}} & MuJoCo & rigid/artic. dyn. & \color{teal}{\checkmark} & \color{teal}{\checkmark} & \color{teal}{\checkmark} &  Disc.\& Cont. & {\!+} & tabletop & \color{teal}{\checkmark}\\
        \bottomrule
    \end{tabular}
    \label{tab:simulators}
    \vspace{-1em}
\end{table*}

\noindent{\bf Simulation environments for manipulation} focused on combined visual and language robotic tasks.
State-of-the-art simulators ManipulaTHOR~\cite{Ehsani2021ManipulaTHOR:Manipulation} (extension of AI2-THOR~\cite{Kolve2017AI2-THOR:AI}), CoppeliaSim~\cite{coppeliaSim}, or iGibson~\cite{li2022igibson}  provide  non-photorealistic rendering, lacking accurate shades and reflections. This is in contrast to Blender, which can generate photorealistic scenes with procedural object models. ISAACSim~\cite{Makoviychuk2021IsaacLearning} is a resource-intensive, closed-source simulator with limited customisability. ThreeDWorld~\cite{Gan2021ThreeDWorld:Simulation} also uses a commercial renderer and does not provide the infrastructure for robot manipulation tasks, such as motion planners. Unlike prior works, we pay particular attention to conditional action planning and execution in robot manipulation while preserving accurate physics modelling between the bodies. We use MuJoCo, which is faster and offers higher physics accuracy than Unity and generally outperforms PyBullet and PhysX in robot manipulation tasks, especially in simulations involving articulated systems with many joints or connected elements~\cite{erez2015simulation}. While Unity is primarily designed for gaming, MuJoCo is optimized for reinforcement learning~\cite{kaup2024review} and robot arm control. The PyBullet simulator was started as a more accessible open-source alternative and offers domain randomization capabilities to enable sim-to-real transfer. \nsap{}, with its procedural scene generation, allows the creation of diverse training data without sacrificing the photorealistic image quality. \nsap{} also offers both continuous and discrete action spaces, in contrast to only discrete action spaces such as in ThreeDWorld~\cite{Gan2021ThreeDWorld:Simulation} and ManipulaThor~\cite{Ehsani2021ManipulaTHOR:Manipulation}, or pure continuous control of the robot as n Robosuite~\cite{zhu2020robosuite}. For example, ManipulaThor~\cite{Ehsani2021ManipulaTHOR:Manipulation} considers grasping as a discrete action, while we enable the execution of sequences of various custom primitive actions (e.g., approach, gripper closing, lifting, etc.) or continuous control of the robot. From these simulators, only iGibson offers non-kinematic continuous and discrete object states \eg temperature or burned. Our {\bf MuBlE environment} differs from the aforementioned manipulator simulators by a unique and open-source combination of 1) high-quality visual data and realistic physics modelling, which is a key advantage when attempting sim2real transfers (as shown in our experiments), 2) the ability to select between operating on continuous action space or the usage of primitive actions, 3) considering both visible and non-visible continuous object states (e.g., weight, or stiffness) and providing a set of primitive action controllers for these, and 4) supporting seamless integration of closed-loop reasoning methods by providing high-speed physics and keyframe action-render loops. We present a comparison between different simulators in Table~\ref{tab:simulators}.

\noindent{\bf Benchmarks for language-conditioned manipulation.}
The advantages of the MuBlE environment enabled us to prepare a unique {\bf benchmark for closed-loop embodied reasoning (SHOP-VRB2)} that extends SHOP-VRB~\cite{Nazarczuk2020SHOP-VRB:Perception}. SHOP-VRB is a dataset closely related to the robot manipulation scenarios considered in this work. It features various objects suitable for robot manipulation that can be easily obtained. There are many high-quality 3D models of various instances of these objects, which make them suitable for generating synthetic scenes via Blender rendering that includes procedurally generated materials rather than simply texturing the mesh. We, therefore, extend it to a dataset and a benchmark SHOP-VRB2 that enables the development and testing of systems capable of reasoning on non-visual attributes and performing interactive tasks. Compared to other benchmarks, such as ALFRED~\cite{Shridhar2020ALFREDTasks} (based on RoboTHOR), a benchmark for understanding natural language instructions in robotics, VLMbench~\cite{zheng2022vlmbench}, based on RLbench~\cite{james2020rlbench} and CoppeliaSim~\cite{coppeliaSim}, which adds linguistic commands and realises automatic complex tasks builders, IKEA Furniture Assembly~\cite{Lee2021IKEATasks}, which provides a benchmark for various assembly tasks and CALVIN benchmark~\cite{mees2022calvin} offering natural language instructions for long-horizon manipulation tasks, our SHOP-VRB2 benchmark requires long-horizon closed-loop embodied reasoning, that involves manipulation of the world to acquire knowledge about non-visual object properties in order to complete the task or answer a query.

%% file: Sections/environment.tex
\section{{MuBlE} Environment modules}
\label{sec:environment}

\begin{figure}
\centering
    \centering
     \includegraphics[width=0.9\linewidth]{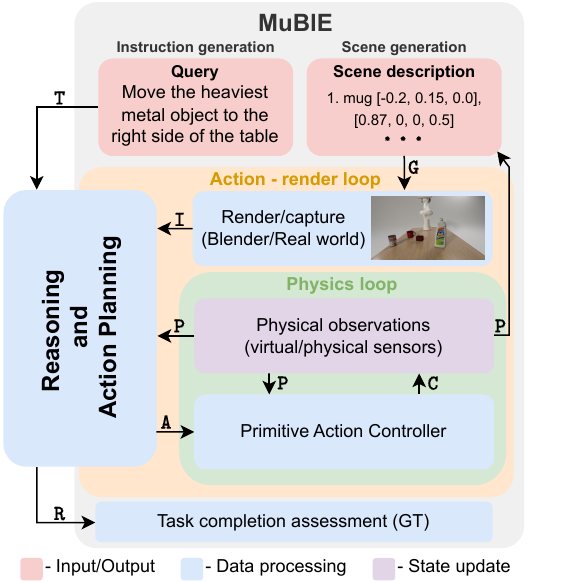}
    \caption{
    A diagram showing individual modules of the \nsap{} environment, including how a reasoning approach might be integrated within the \nsap{} environment. Example instruction and scene from \dataset{} benchmark are shown. Symbols for transferred data: $\mathtt{T}$ - query text, $\mathtt{I}$ - image, $\mathtt{G}$ - scene graph, $\mathtt{P}$ - physical observations, $\mathtt{C}$ - control signal, $\mathtt{A}$ - primitive action to take, $\mathtt{R}$ - returned result, $\mathtt{GT}$ - ground truth data.
    }
    \label{fig:env_diag}
    \vspace{-1em}
\end{figure}

This section presents our \nsap{} environment for simulating manipulation tasks. MuBlE is an interactive environment consisting of MuJoCo and Blender-based modules, capable of physics simulation coupled with high-quality image rendering.  The main modules and data flow are presented in Fig.~\ref{fig:env_diag}. \nsap{} is built on $\mathtt{robosuite}$~\cite{zhu2020robosuite}, which is a simulation framework suitable for creating robotic environments inside the physics engine MuJoCo. We equip our environment with high-quality rendering powered by Blender. It is designed for a generic tabletop scenario with a single robotic manipulator and a gripper. We use the same set of robots and grippers as $\mathtt{robosuite}$. We provide a set of object models in MuJoCo and their counterparts in Blender, along with templates to create new items.

We first introduce the underlying data representation in MuBle, then the Action loop of the proposed environment, followed by details of the Physics loop and capabilities of MuBle in terms of dataset and benchmark generation.

\subsection{Data representation}

MuBlE initialises the scene based on the given specification and records a scene graph that is exposed to both MuJoCo and Blender. MuJoCo representation is updated in response to the control signal from the Reasoning and Action Planning module. This is followed by an update in the scene graph, which propagates to the scene built in Blender. MuBlE exposes an interface for rendering the current, accurate scene image at any time.

\noindent {\bf Scene graph.} MuBlE uses a scene graph as the underlying representation of the scene. The scene graph encodes positions, orientations, and measurements for all objects collected in the environment. It is used to enforce consistency between MuJoCo and Blender scenes. The initial scene graph can be generated using our Scene generator (see Sec.~\ref{s:tools}).

\subsection{Scene capture/render}
The real scene can be captured by a camera, or a synthetic scene can be rendered with Blender. A keyframe is rendered after a motion corresponding to a primitive action is finished, \eg moving from object 1 to object 2, approaching grasping pose, closing the gripper, etc. We have chosen the keyframe-based approach as it allows us to use Blender for rendering frames, which have sufficient quality for sim2real transfer and sufficient speed to allow for recovering from failures with the closed-loop approach. The high-quality procedural rendering allows us to visually represent various types of properties (\eg color, material, texture) under different shading and lighting conditions.

Note that high-quality rendering of every frame can be used for training, but it is too slow for real-time experiments. However, many tasks are achievable with low-rate visual reasoning and fast feedback loops from other sources of information (\eg force control). Example keyframes rendered during the execution of a task are presented in Fig.~\ref{fig:test}~(Left).

\subsection{Action loop and connection to action planner}

\nsap{} is designed to operate in two levels of interaction with the environment; we denote it as an Action loop and a Physics loop. The former allows for interaction with the environment in response to visual cues, whereas the latter allows for precise control of the robot. This design facilitates the development of methods, including high-level plans and robot control in a closed loop.

The Action loop interacts with the reasoning approach through visual outputs and primitive action inputs. This allows the corresponding approach to respond to visual changes in the environment. To this end, \nsap{} awaits a primitive action $\mathtt{A}$ from the Action Planning module as well as the predicted state of the scene, including positions and orientations of all the objects. Every primitive action generates a desired trajectory for the robot end effector (the gripper). MuBle provides a trajectory planning module with obstacle avoidance and exposes an interface for a custom Motion Planner. When the trajectory for the given primitive action $\mathtt{A}$ is achieved, the new visual observation is automatically rendered.

\begin{figure}
    \centering
    \includegraphics[width=0.92\linewidth]{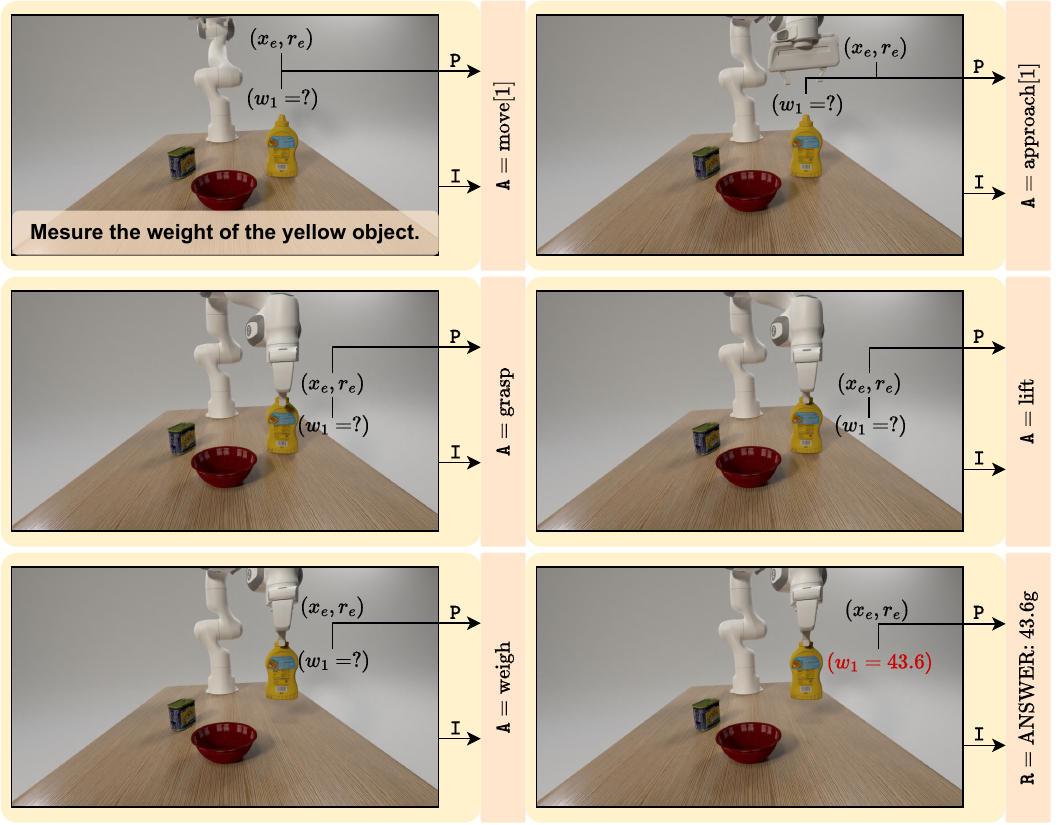}
    \caption{
    An example of the interaction between \nsap{} (in yellow) and a reasoning method (in orange). Figure presents selected measurements $\mathtt{P}$ and primitive actions $\mathtt{A}$ generated based on them, followed by a corresponding update of the scene in the environment.
    }
    \label{fig:pipeline}
\vspace{-1em}
\end{figure}

Our environment allows easy interfacing with an action planner. Such a reasoner can be integrated as a separate module within our environment. Its goal is to predict the next primitive action $\mathtt{A}$ and its target, given the current scene graph $\mathtt{G}$ and subgoal $\mathtt{S}$ (see Fig.~\ref{fig:env_diag} and Fig.~\ref{fig:pipeline}).

\subsection{Physics loop}
\label{s:physics}
The physics interactions are calculated with the use of MuJoCo engine as shown in Fig.~\ref{fig:env_diag}. With a user-defined timestep, the control signal $\mathtt{C}$ is applied to the manipulator, and the corresponding forces are calculated and applied to all objects in the scene. Similarly to $\mathtt{robosuite}$, the control signal takes the form of the displacement of the end effector, \ie position and orientation change, along with a binary signal for opening and closing of the gripper. By default, the control signal is translated to desired joint forces by the Operational Space Controller \cite{Khatib1987AFormulation}, which can be customised. A control signal to the environment can be provided by the user directly or can be obtained from the primitive action controller, which calculates the error of the current end effector pose and the planned trajectory.

Every step of applying the control $\mathtt{C}$ affects the physics of the scene and generates a set of observations $\mathtt{P}$, which include the position and orientation of the end effector, joint positions and velocities, status of the gripper closure. Observations of measured non-visual properties, such as stiffness, weight, or elasticity, are also included.  Additionally, measurements related to all objects in the scene may be collected, such as positions, orientations, bounding boxes, and indication of whether the object is currently in the gripper. 
 Any custom sensor can be added to \nsap as in $\mathtt{robosuite}$.

\subsection{Primitive actions}
\label{s:action}
To enable the planning of various manipulation tasks, we design a set of primitive actions $\mathtt{A}$ that can be easily extended in \nsap{}:
\begin{itemize}
    \item $\mathtt{move}$ -- moves the end effector towards a given target (\eg another object or part of the table),
    \item $\mathtt{approach}$ -- positions the end effector in a grasping position with respect to the target object
    \item $\mathtt{close\_ gripper}$, $\mathtt{open\_ gripper}$ -- to grasp or release,
    \item $\mathtt{lift}$, $\mathtt{lower}$ -- lifts or lowers the end effector,
    \item $\mathtt{weigh}$ -- weighs the object in the gripper,
    \item $\mathtt{squeeze}$ -- squeezes the object in the gripper to measure its stiffness.
\end{itemize}

The primitive action controller implements a control that can execute action $\mathtt{A}$ on target object in MuJoCo physics engine or real robot. The physics loop, discussed in more detail in Sec.~\ref{s:physics}), calculates physics and collects observations $\mathtt{P}$ (\eg pose of end effector, physical measurement, force in the gripper, etc.).  A control signal is generated in every step of the physics loop until the path is completed. After the execution, a new key frame is captured in a real-world setup or rendered in simulation.

The freedom to implement primitive action controllers with access to the simulation physics loop allows for the implementation of high-level (discrete) actions (\eg $\mathtt{approach}$) and low-level (continuous) actions (\eg $\mathtt{move\_vel}$ that move with a given velocity for a given time). The continuous actions require the agent to parameterise a primitive action controller with real valued parameters, while discrete actions can be selected from a finite amount of options.

Primitive action execution with obstacle avoidance requires the positions and orientations $\mathtt{P}$ of the objects in the scene. Every primitive action, including gripper closure, generates a desired trajectory (position and orientation) $\mathtt{P}$ for the end effector.  The action of approaching the grasp position entails a set of pre-coded grasp sequences. We implement grasping of cylindrical objects (depending on their position and size), grasping by the edge, grasping by the handle, etc. We use a simplified model, as simulated grasping is often not reliable and is inconsistent with real grasping. Whenever grasping pads on all gripper fingers are in contact with the same object, it is considered as grasped. Thereafter, the object's pose is fixed with respect to the end effector. Further, whenever the gripper starts to open, the object is detached from the gripper body with an initial velocity of 0.
{More details can be found on the GitHub page: {\small\url{https://github.com/michaal94/MuBlE}}}

\subsection{Data generation tools}
\label{s:tools}

\noindent{\bf Scene generator}
is a tool in \nsap{} that can procedurally generate data for training models. We provide a template scene that is compatible with the renderer within the framework. The procedural algorithm takes a set of object models, which can be customised, and randomly places them on the tabletop. Desired properties of the objects, such as colour, material, and size, can also be randomised.  We check for collisions between meshes, unlike previous approaches that check only bounding boxes. Afterwards, a maximal desired level of occlusion is ensured.  Finally, we render an image of the scene, generating full ground truth, including segmentation masks for all the objects, the robot, and the tabletop, as well as a depth map. Fig.~\ref{fig:scenes} presents some example scenes generated with \nsap{}. Objects in the scene are described in Sec.~\ref{sec:dataset}.
 
\noindent{\bf Instruction generator}
\label{s:instruction}
is another tool in \nsap{} that, given the scene, generates natural language instructions or tasks that require reasoning and interaction. We propose a template-based algorithm where we improve upon CLEVR and SHOP-VRB, which are time-consuming and scale poorly when more adjectives are introduced in the templates. This is due to evaluating all possible combinations of describing words and validating the descriptions with respect to the given scene. Instead, we implement an efficient rejection mechanism. We first generate all combinations of short descriptions for all scene objects. Short descriptions of objects are more natural than in \cite{Johnson2017CLEVR:Reasoning, Nazarczuk2020SHOP-VRB:Perception}. Next, we evaluate the descriptions using scene constraints \eg the target of the instruction has to satisfy the constraint of being \textit{pickupable}; otherwise, the description is rejected. The instruction targets are randomly selected from the validated descriptors and used to generate a ground truth symbolic program. We automatically produce the ground truth actions by reasoning backwards with predefined pairwise relations \eg lifting an object requires grasping it, grasping requires approaching to the grasping position, etc.). Fig.\,\ref{fig:scenes} presents example instructions generated for the given scenes using templates that are described in Sec.~\ref{sec:dataset}.

%% file: Sections/dataset.tex
\section{\dataset Dataset} \label{sec:dataset}

\begin{figure}[!t]
    \centering
    \begin{minipage}{0.95\linewidth}
    \includegraphics[width=\linewidth]{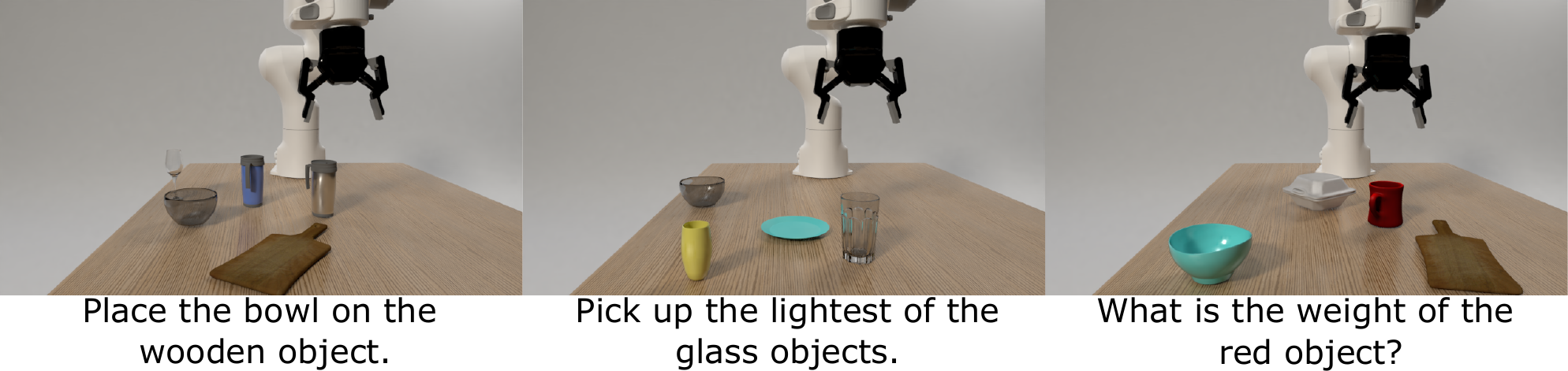}
    \end{minipage}
    \caption{Example of \dataset: example simulated scenes and corresponding instructions in natural language generated with \nsap{} (in the dataset, instructions left to right belong to tasks 7, 3, and 1 in Tab.~\ref{tab:templates}).}
    \label{fig:sample_seq}
    \label{fig:scenes}
    \vspace{-1em}
\end{figure}

We introduce \dataset{} dataset created in \nsap{} for training and benchmarking. The dataset includes a set of scenes with instructions to perform various tasks (\eg~\emph{Stack metal objects from heaviest to lightest}). Tasks are designed to enforce reasoning simultaneously on the visual observations (recognising attributes of objects and their relations) and continuous physical measurements, taking the feedback loop into account. Every example is accompanied by a ground truth sequence of actions for successful execution, along with visual observations and detailed scene graphs. Some examples can be found in Fig.~\ref{fig:sample_seq}. We include weight measurement as a representative example of estimating non-visual object properties through manipulation due to the ease of repeatability for other researchers, but we also demonstrate stiffness measurements. Other properties, such as roughness, are also possible but require another template for training and an implementation of the control. 

\subsection{Benchmark data}
\label{s:benchmark}

\noindent{\bf \dataset scenes} include $12000$ realistically rendered scenes 
split to train, validation, and test sets: $10000$:$1000$:$1000$. The scenes contain typical household objects which are easily accessible. Following \cite{Nazarczuk2020SHOP-VRB:Perception}, various instances of objects are included: baking trays, bowls, chopping boards, food containers, glasses, mugs, plates, soda cans, thermoses, and wine glasses. The scenes are generated with $4$ to $5$ objects per scene (their respective distribution is $47.3\%$ and $52.7\%$). We make sure the randomisation of the materials (plastic, metal, glass, rubber, wood) is realistic. Out of $54317$ placed objects, a subset was selected randomly, with $1330$ instances of the least common (big chopping board model due to many possible collisions), and $3486$ instances of the most common model (bowl -- less prone to collisions).

\noindent{\bf YCB scenes} include 30 simulated benchmarking scenes with 9 YCB-Video \cite{Xiang2018PoseCNN:Scenes} objects and 3 randomly generated scenes for each of the benchmarking tasks described below. 9 YCB objects were selected so that the objects share various types of visual/physical attributes (cleanser, mustard, mug, bowl, tomato can, Cheez-It, sugar box, meat can, foam brick).

\begin{table}
    \caption{\scriptsize Instruction templates corresponding to the benchmarking tasks in the proposed dataset.}
    \begin{tabularx}{\linewidth}{rl}
    \toprule
    No. & Instruction Templates  \\
    \midrule
        1. & Measure the weight of the $\mathtt{OBJ1}$. \\
        2. & What is the weight of all $\mathtt{OBJ1}$s? \\
        3. & Pick up the $\mathtt{WS1}$ of all $\mathtt{OBJ1}$s. \\
        4. & Place the $\mathtt{OBJ1}$ on the $\mathtt{TP1}$ part of the table.\\
        5. & Remove all $\mathtt{OBJ1}$s from the $\mathtt{TP1}$ part of the table. \\
        6. & Place the $\mathtt{WS1}$ of all $\mathtt{OBJ1}$s on the $\mathtt{TP1}$ part of the table. \\
        7. & Stack the $\mathtt{OBJ1}$ on top of the $\mathtt{OBJ2}$. \\
        8. & Place the $\mathtt{WS1}$ of all $\mathtt{OBJ1}$s on top of the $\mathtt{OBJ2}$. \\
        9. & Stack the $\mathtt{OBJ1}$ on top of the $\mathtt{OBJ2}$ on top of the $\mathtt{OBJ3}$. \\
        10. & Stack all $\mathtt{OBJ1}$s from heaviest to lightest. \\
    \bottomrule
    \end{tabularx}
    \label{tab:templates}
\end{table}

\noindent{\bf Benchmarking tasks} \label{sec:instructions} We assign one instruction to each scene to introduce more diversity in visual observations. We designed 10 classes of benchmarking tasks revolving around moving and stacking objects based on their visual (colour, material, shape) and physical properties (weight). Example templates are presented in Tab.~\ref{tab:templates}.
The task types are closely related to the instruction templates in Tab.~\ref{tab:templates} and involve 1) measuring weight of a single object, 2) measuring weight of multiple objects, 3) picking up based on weight, 4) moving single object 5) moving multiple objects 6) moving based on weight 7) stacking objects 8) stacking objects according to weight 9) stacking 3 objects 10) ordering objects according to their weight. $\mathtt{OBJx}$ refers to a description of an object consisting of a set of visual properties (chosen randomly and validated), \eg \textit{the red object} presented in Fig.~\ref{fig:teaser}. Note that tasks may refer to either one specific unique object or a set of objects sharing a certain property \eg template 1 and 2 in Tab.~\ref{tab:templates}. Further, $\mathtt{TPx}$ specifies a part of the table, \eg the left and the right part. Finally, $\mathtt{WSx}$ refers to the weight specifier, \ie distinction whether the lightest or the heaviest object from the set is the target of the instruction. The length of instructions ranges between $5$ and $16$ words. The resulting sequences contain between $5$ (measure the weight of the single object) and $46$ (stacking several items according to weight) primitive actions. All instructions are accompanied by symbolic programs in CLEVR-IEP \cite{Johnson2017InferringReasoning} format and task specifications with respect to the scene graph.

\noindent{\bf Ground truth} for visual and physical observations are provided for every scene and its benchmarking instruction. An observation is taken after every primitive action is executed and contains: position and orientation of the end effector, gripper status, weight measurement, ground truth scene graph, segmentation masks, and action with its target.

\noindent{\bf Benchmarking metrics}
The suggested metrics for benchmarking reasoning methods are the rate of successful task execution (\eg stacking) and correct question answer (\eg weight query). The reported accuracies should also be split into task types presented in Tab.~\ref{tab:templates}.

%% file: Sections/baseline.tex
\section{Experiments}
\label{sec:results}
\begin{figure*}
    \centering
    \includegraphics[width=0.94\linewidth]{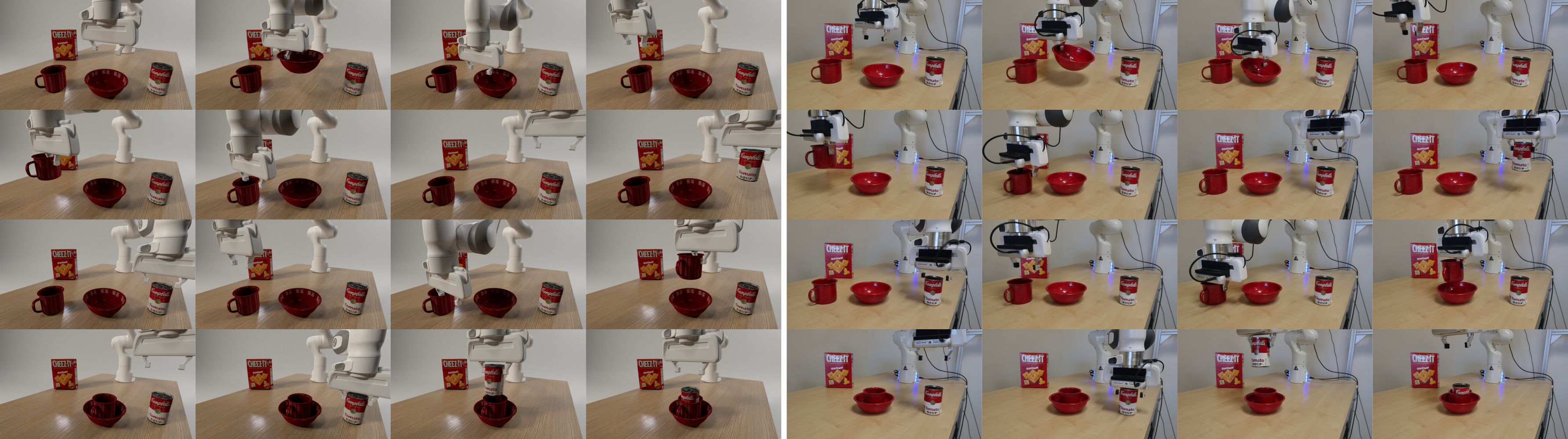}
    \caption{Examples of visual observations (selected frames) generated for the actions corresponding to the instruction: \textit{Stack metal objects from heaviest to lightest}. (Left) Simulated YCB scenes rendered by Blender in \nsap and (Right) corresponding real YCB scenes captured by Realsense camera during the real-world experiment using the reasoning pretrained in the \nsap environment on the simulated \dataset dataset.
    }
    \label{fig:test}
    \vspace{-1em}
\end{figure*}

To showcase the capabilities of the simulator and the proposed benchmark, we employ a baseline action planner~\cite{Nazarczuk2025}. We show the results for two datasets: 1) in simulation for \dataset{} dataset (Sec.~\ref{sec:benchmark_shop}), and 2) on a set of 30 benchmarking scenes in simulation and in the real world with YCB objects~(Sec.~\ref{sec:benchmark_ycb}). 

\noindent{\bf{Baseline reasoning method.}} As a baseline method we evaluate CLIER method~\cite{Nazarczuk2025} - it employs a transformer as an action planner. The transformer predicts the next primitive action $\mathtt{A}$ and its target, given the graph of the current scene $\mathtt{G}$ and the sub-goal $\mathtt{S}$. ResNet is used to process visual inputs, and a Seq2Seq networkis used  to process the textual inputs. The baseline method is integrated in \nsap{}. 

\begin{table}
    \caption{ 
    Success rates for the Baseline method on \dataset{} (sim) and YCB dataset (sim/real).
    }
    \label{tab:subtask}
\centering
\scriptsize
    \begin{tabularx}{0.85\linewidth}{lYYY}
    \toprule
    Success [$\%$] & \dataset{} & \multicolumn{2}{c}{YCB\,\,\,} \\
    Task type & Sim & Sim & Real  \\
    \midrule
    Weight single & $74.0$ & $66.7$ & $88.9$  \\
    Weight multi &  $65.0$ & $100$ & $66.7$ \\
    Pick up weight & $49.0$  & $100$ & $88.9$  \\
    Move single &  $76.0$ & $66.7$ & $100$ \\
    Move multi & $47.0$  & $100$ & $44.4$  \\
    Move weight & $23.0$  & $100$ & $100$ \\
    Stack & $56.0$  & $66.7$ & $66.7$ \\
    Stack weight & $31.0$  & $33.3$ & $22.2$  \\
    Stack three & $0.0$  & $66.7$ & $0.0$  \\
    Order weight & $18.0$  & $66.7$ & $100$\\
    \midrule
    Overall & $43.9$ & $76.7$ & $64.4$ \\
    \bottomrule
    \end{tabularx}
\vspace{-1em}
\end{table}

\subsection{\dataset experiments}
\label{sec:benchmark_shop}

Results for the baseline method are presented in Table \ref{tab:subtask}, which shows success rates for individual tasks.  We observe high success rates (100\%) for the tasks that include manipulation of single objects (weighing or moving one object), and a strong decrease in accuracy for multi-object manipulation (stacking more objects), with the lowest accuracy (33\%) for multi-step manipulation that requires embodied reasoning (Stack weight: stacking objects based on their weight), highlighting the struggles of current reasoning methods to perform well on long-horizon tasks. 

Further, we identify the most common reasons for failure: 1) Execution error ($14.4\%$) accounts for failures in the execution of primitive actions which may arise from inaccurate scene description (\eg typically object pose), 2) Scene inconsistency ($12.6\%$) refers to mistakes in tracking object IDs between frames, and 3) Loop detection ($10.8\%$) arises when primitive action chains are repeated (\eg when approaching an object to grasp with misaligned position).

Finally, we believe that the overall accuracy of \textbf{$43.9\%$} on \dataset{} demonstrates the complexity of the benchmark, as models struggle to achieve high performance. This relatively low accuracy suggests that the dataset forms a challenging benchmark for visual and interactive reasoning, making it a valuable tool for evaluating and improving future methods.

\subsection{Real-world experiments}
\label{sec:benchmark_ycb}

In this section, we report comparative results for real YCB scenes that mirror 30 YCB scenes introduced in Sec.~\ref{s:benchmark}. In Fig.~\ref{fig:test}\,(Left) we show an example of the execution of the task using the simulated environment and in Fig.~\ref{fig:test}\,(Right) we show an example of the corresponding execution in the real environment.

\begin{figure}
    \centering
    \includegraphics[width=\linewidth]{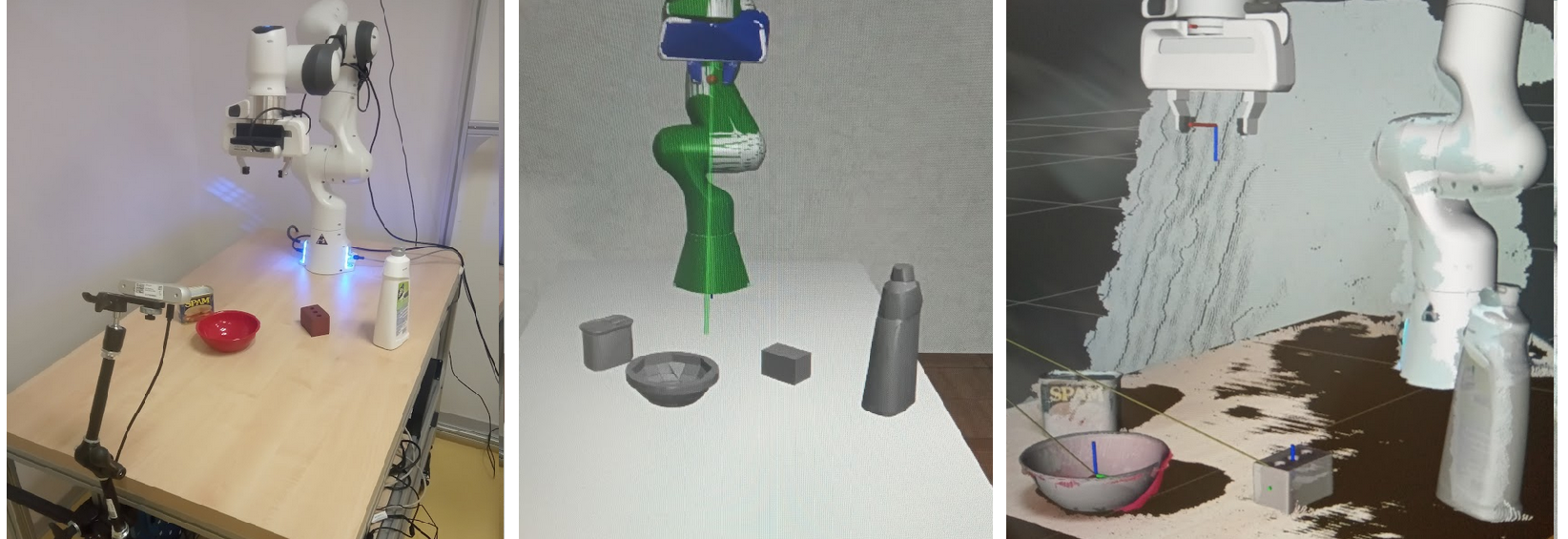}
    \caption{The real setup with YCB objects (left), corresponding MuJoCo simulation using estimated poses (middle), and RViz visualisation of colored pointcloud with overlaid gray models detected by CosyPose (right).}
    \label{fig:real_robot}
    \vspace{-1em}
\end{figure}

\noindent{\bf Experimental Setup} includes Franka Emika Panda~\cite{franka} arm with $7$ degrees of freedom and a $2$-finger parallel gripper. An (extrinsically calibrated) Intel Realsense D455 camera is set to face the robot and capture the front view of the scene (see Fig.~\ref{fig:real_robot}). To allow the control of the real robot, we developed a thin ZMQ~\cite{zeromq} based communication layer that emulates the same interface as for the simulated robot (\ie, motion generation using position-based servoing). Special purpose skills include measuring stiffness via squeezing the object with different forces and measuring deformation or weight via lifting and joint torque differences. Note that the weighing has a relative error of less than $\pm 10\%$ in the range $0-200$~g compared to the ground truth weight. The stiffness measurement has a coefficient of variation ranging from $1.6-3.4$~\%, which is highly repeatable. However, we were not able to obtain ground truth data for stiffness.

\noindent{\bf Sim2Real} Instead of the rendered images, we used raw RGB images as an input to the baseline reasoning method, which was trained exclusively in the simulator and remained unchanged for both YCB simulation and real-world experiments. The same reasoning model is evaluated in 30 synthetic and real scenes. Since the reasoning modules operate on the scene graph, they are agnostic to visual input. To construct the scene graph, we used CosyPose~\cite{Labbe2020CosyPose:Estimation} to extract object poses and the same ResNet as in the simulation to infer visual object properties like material, texture or color from the RGB images. The system achieved 76.7\% accuracy in simulation and 64.4\% on the real dataset, highlighting the importance of the high-quality renderer for the sim-to-real transfer. When trained using the default $\mathtt{robosuite}$ renderer, the accuracy on the real dataset dropped to nearly zero, as the model failed to correctly detect both objects and their visual properties in real-world scenes.

%% file: Sections/conclusions.tex
\section{Discussion and Conclusions} \label{sec:conclusions}

In this paper, we presented the \nsap{} environment (built on top of the $\mathtt{robosuite}$ simulation environment) that incorporates MuJoCo physics simulation with a high-quality renderer and enables the generation of multi-modal demonstration data for robot manipulation tasks. By introducing a high-quality renderer into the simulation pipeline, our system overcomes the sim2real gap and, therefore, becomes a useful, efficient and low-cost alternative to collecting data from robotic arm demonstrations in the real world. In our framework, the presence of shadows, varying lighting, and reflections poses a significant visual challenge for planning systems, similar as real-world images, making it both a challenging benchmark and a useful data source for sim2real transfer. Our fully modular environment enables both data generation and benchmarking of simultaneous reasoning in visual and physical space. The results from simulated and real-world experiments showed the ability to successfully transfer between the simulated and real environment, emphasising the usefulness of MuBlE in generating training data for transfer learning.

This work provides a new challenge and a benchmark for future systems that aim at bridging the gap between physics simulation for robotic manipulators, realistic visual simulations and execution in real environment.